%% file: main.tex
\documentclass[letterpaper, 10 pt, conference]{ieeeconf}  % Comment this line out if you need a4paper
\IEEEoverridecommandlockouts                              % This command is only
\usepackage{xspace}
\usepackage{colortbl} 
\usepackage{multirow}
\usepackage{biblatex}
\usepackage{bm}
\usepackage{float} 
\usepackage{amsfonts}
\usepackage{hyperref}
\usepackage{booktabs}
\usepackage{graphicx} 
\usepackage{caption}
\usepackage{lipsum}
\usepackage[usenames,dvipsnames,table]{xcolor}
\hypersetup{
    colorlinks=true,
    linkcolor=MidnightBlue,
    filecolor=magenta,      
    urlcolor=MidnightBlue,
    citecolor=MidnightBlue,
} 
\newcommand{\model}[0]{DSPv2\xspace}

\definecolor{perfcolor}{HTML}{008000}
\newcommand{\drop}[1]{\ensuremath{#1\!\rightarrow}}
\newcommand{\customfootnotetext}[2]{{% Group to localize change to footnote
  \renewcommand{\thefootnote}{#1}% Update footnote counter representation
  \footnotetext[0]{#2}}}% Print footnote text

\title{\LARGE \bf
DSPv2:  Improved Dense Policy for Effective and Generalizable \\ Whole-body Mobile Manipulation
}
\author{Yue Su$^{1,2,3}$~ Chubin Zhang$^{2,4}$~ Sijin Chen$^{1}$~ Liufan Tan$^{2}$~ Yansong Tang$^{4}$~ Jianan Wang$^{2\ddagger}$~ Xihui Liu$^{1\dagger}$ \\
{\small {$^{1}$The University of Hong Kong}~
{$^{2}$Astribot}~
{$^{3}$Xidian University}~
{$^{4}$Tsinghua University}}
}

\begin{document}
\makeatletter
\let\@oldmaketitle\@maketitle% Store \@maketitle
\renewcommand{\@maketitle}{\@oldmaketitle% Update \@maketitle to insert...
% \vspace{-0.1cm}
\centering
\includegraphics[width=\linewidth]{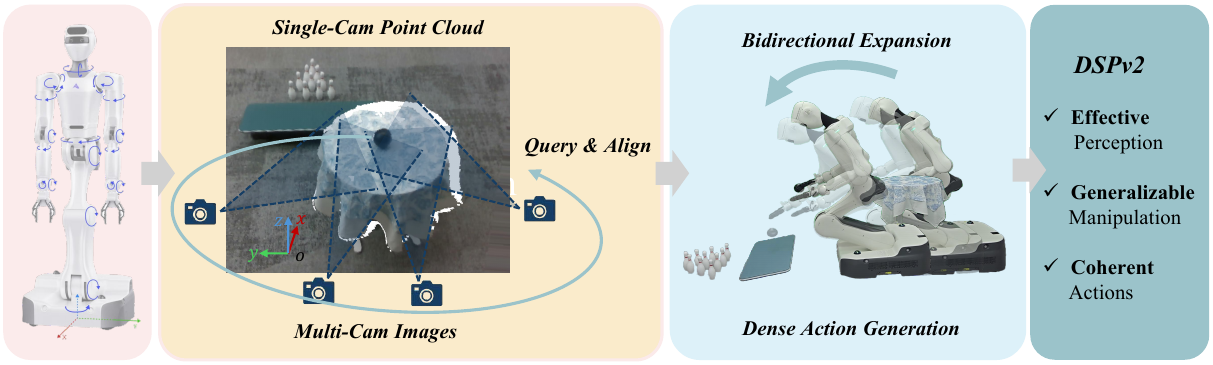}
\vspace{-0.2cm}
\captionof{figure}{\textbf{\model is a whole-body mobile manipulation policy} that achieves generalizable performance by fusing multi-view 2D semantic perception with 3D spatial awareness, and generates coherent whole-body actions via dense action head.}
\label{fig:teaser}
}
\makeatother
\maketitle

\thispagestyle{empty}
\pagestyle{empty}
\addtocounter{figure}{-1}

\customfootnotetext{$\ddagger$}{Project Lead}
\customfootnotetext{$\dagger$}{Corresponding Author}

\input{sec/0_abstract}

\input{sec/1_intro}
\input{sec/2_related}
\input{sec/3_method}
\input{sec/4_exp}
\input{sec/5_limitations}
\input{sec/6_conclusion}
% \input{sec/7_ack}
% \clearpage

\renewcommand*{\bibfont}{\footnotesize}
\printbibliography 
\end{document}

%% file: sec/0_abstract.tex
\begin{abstract}
Learning whole-body mobile manipulation via imitation is essential for generalizing robotic skills to diverse environments and complex tasks. However, this goal is hindered by significant challenges, particularly in effectively processing complex observation, achieving robust generalization, and generating coherent actions. To address these issues, we propose \model, a novel policy architecture. \model introduces an effective encoding scheme that aligns 3D spatial features with multi-view 2D semantic features. This fusion enables the policy to achieve broad generalization while retaining the fine-grained perception necessary for precise control. Furthermore, we extend the Dense Policy paradigm to the whole-body mobile manipulation domain, demonstrating its effectiveness in generating coherent and precise actions for the whole-body robotic platform. Extensive experiments show that our method significantly outperforms existing approaches in both task performance and generalization ability. Project page is available at: \url{https://selen-suyue.github.io/DSPv2Net/}. 
\end{abstract}

%% file: sec/1_intro.tex
\section{Introduction}
\label{sec:intro}

Recent advances in imitation learning have enabled robust performance on manipulation tasks using single or dual-arm configurations~\cite{ACT,DP,DP3}, culminating in the development of generalizable~\cite{cage,airexo2} and general~\cite{gr3,octo,openvla} policies. In stark contrast, applying these techniques to whole-body household robots, which possess immense market potential and real-world applicability, still faces considerable difficulties, even in achieving reliable task-specific downstream policies~\cite{brs}.

This difficulty stems primarily from two intertwined aspects: the complexity of the robot's morphology and the expanded scope of its task environment.
From the robot's perspective, a whole-body household robot possesses significantly higher DoF and a more expansive workspace compared to fixed-base robot arms~\cite{duocorewb}. Tasks executed by such a robot often necessitate the coordination of dual-arm manipulation, whole-body motion, and chassis navigation. This coupling results in a complex and high-dimensional action space and imposes stringent requirements on the coordination among its joints.
From the scene's perspective, a majority of existing policies~\cite{ACT,RISE} are developed for constrained, tabletop manipulation tasks. In contrast, whole-body robots are expected to manipulate in a broad spectrum of domestic and outdoor environments. Thus, they often require multiple cameras to achieve comprehensive awareness. These complicated scene and observation lead to a diversely distributed observation space, thereby substantially compounding the difficulty of policy learning~\cite{dsp}.

Research addressing this problem is scarce. Current solutions~\cite{brs} obtain observations by projecting and merging multi-view point clouds into the base frame, followed by a sampling step. Actions are then generated with diffusion head. However, the sparse sampling required by the 3D vision backbone~\cite{pointnet} in this method limits the utilization of multi-view observations, which degrades task execution accuracy. Furthermore, this approach fails to address the problem of generalization. This is, however, crucial for the policy's practical applicability. Beyond the aforementioned issues, the coupled dynamics of whole-body systems present another major hurdle. Minor misalignment between components, such as the manipulator and the mobile base, can rapidly amplify, thereby undermining the coherence of the entire motion~\cite{brs,acdit}.

To address these challenges, we propose \textbf{\model}, an improved Dense Policy~\cite{dsp} for whole-body mobile manipulation. As shown in Figure~\ref{fig:teaser}, our model is designed to effectively leverage multi-view observations to achieve robust whole-body manipulation performance, generate coherent actions by bidirectional autoregressive way, while also generalizing to unseen objects and environments. 

% Specifically, we design an effective method for multi-view learning and 2D-3D feature fusion. First, \model projects the uncolored point cloud from the head camera into the base frame as a global observation. This point cloud is then processed by a Sparse 3D Encoder~\cite{spatioenc} to extract voxel-level spatial features. Next, RGB images from cameras on the head, wrist, and torso are encoded into patch-level semantic features using a fine-tuned lightweight vision foundation model~\cite{dinov2}. Subsequently, we design a Q-former~\cite{blip2} that uses the positional information of the spatial voxels to query for corresponding semantic features within the multi-view 2D feature maps. These semantic features are then aligned with the corresponding spatial features to achieve 2D-3D fusion. This decoupled approach allows the spatial features to be processed independently, with the foundation model learning color features to achieve scene-level generalization, while the fusion process facilitates the utilization of geometry-aware multi-view features. For the subsequent action generation, the fused observation features are fed into a Dense Head~\cite{dsp} for coarse-to-fine autoregressive generation. By applying bidirectional attention across the temporal dimension, we mitigate error amplification and incoordination among different robot components during trajectory generation, achieving precise and robust action prediction.

Specifically, to address the challenges of leveraging complex observations for generalizable perception, we design an effective 2D-3D feature fusion method. First, \model projects the uncolored point cloud from the head camera into the base frame as a global observation, which is then processed by a Sparse 3D Encoder~\cite{spatioenc} to extract voxel-level spatial features. Concurrently, RGB images from cameras on the head, wrist, and torso are encoded into patch-level semantic features using a fine-tuned lightweight vision foundation model~\cite{dinov2}. To fuse these modalities, we design a Q-former~\cite{blip2} that uses the positional information of the spatial voxels to query for and align with corresponding semantic features within the multi-view 2D feature maps. This decoupled approach allows the spatial features to be processed independently, with the foundation model learning color features to achieve scene-level generalization, while the fusion process facilitates the utilization of geometry-aware multi-view features.
To tackle the problem of error amplification and motion incoordination in action generation, the fused observation features are fed into a Dense Head~\cite{dsp}. It performs coarse-to-fine autoregressive generation and applies bidirectional attention across the temporal dimension. This mechanism allows us to mitigate error amplification and incoordination among different robot components during trajectory generation, thereby achieving precise and robust action prediction.

We conduct extensive real-world experiments, including five different types of tasks and five distinct generalization tests. We compare our method against widely-used 2D and 3D policies~\cite{DP,DP3}, as well as other whole-body mobile manipulation policies~\cite{brs}. We also apply \model to various action heads and compare the results. Furthermore, we evaluate the effectiveness and generalization of different configurations of our visual encoder. Our key contributions are as follows:
\begin{itemize}
    \item We propose an effective whole-body mobile manipulation policy that fully utilizes multi-view observations by aligning 3D spatial features and 2D semantic features, surpassing existing whole-body policies.
    \item We propose the first generalization approach for whole-body downstream policies and present its strong generalization even with limited task-specific data. 
    \item We extend the Dense Policy to whole-body mobile manipulation and demonstrate its effectiveness in generating coherent and precise whole-body actions.
\end{itemize}

%% file: sec/2_related.tex
\section{Related Work}
\label{sec:related}

\subsection{Imitation Learning for Manipulation}
Advances in imitation learning~\cite{bc,BCZ,IBC} have led to progress in robotic manipulation. For tasks set in complex scenes, 2D policies~\cite{DP,ACT} often learn by encoding multi-view observations to gain a comprehensive understanding. Early 3D policies~\cite{DP3,idp3} encoded downsampled sparse point clouds to obtain robust feature representations after pooling~\cite{pointnet}, a step necessary for stable, denoising-based action generation. However, this method sacrifices comprehensive spatial information, thereby limiting performance~\cite{RISE}. Some 3D foundation policies~\cite{fp3,lift3d} attempt to improve performance through pre-training on a large number of heterogeneous robot datasets, while other approaches use RGB-D representations to mitigate this deficit~\cite{h3dp}. Currently, high-performing 3D policies~\cite{RISE,airexo2,mba} employ sparse convolutions on voxels to extract voxel-level features without losing scene information. This representation balances information preservation with the robustness of feature representation, resulting in better action generation.
We believe this encoding method is well-suited for household scene understanding, especially under complex action and observation space.

Furthermore, in the action generation module, some policies employ generative models to map observations to action distributions~\cite{DP,pi0}. An advantage of these models is their ability to fit diverse action patterns without deterministic sampling~\cite{ACT}. More recently, many studies have explored using autoregressive models~\cite{BERT,GPT} for action generation~\cite{arp,icrt,CARP,h3dp,dsp}. These models leverage temporal dependencies between actions to produce more coherent trajectories and can mitigate the error accumulation that diffusion models may suffer from due to environmental disturbances and domain shift~\cite{d3pm,dsp}. 

\subsection{Policy-level Generalization}
Aside from Vision-Language-Action (VLA) models~\cite{pi0,pi05,openvla}, which possess generalization capabilities from pre-training on large, heterogeneous data~\cite{oxe}, the generalization of policies~\cite{cage,uad} trained on task-specific data often relies on incorporating lightweight vision foundation models~\cite{dinov2,siglip}. This allows them to adapt to variations in scenes and manipulated objects. Current methods~\cite{airexo2} extending this approach to 3D involve processing single-view point cloud and corresponding image independently. The features from the vision foundation model are then projected to their corresponding locations in the point cloud to enable generalizable dual-arm tabletop manipulation.
\begin{figure*}[!t]
  \centering
    \includegraphics[width=\textwidth]{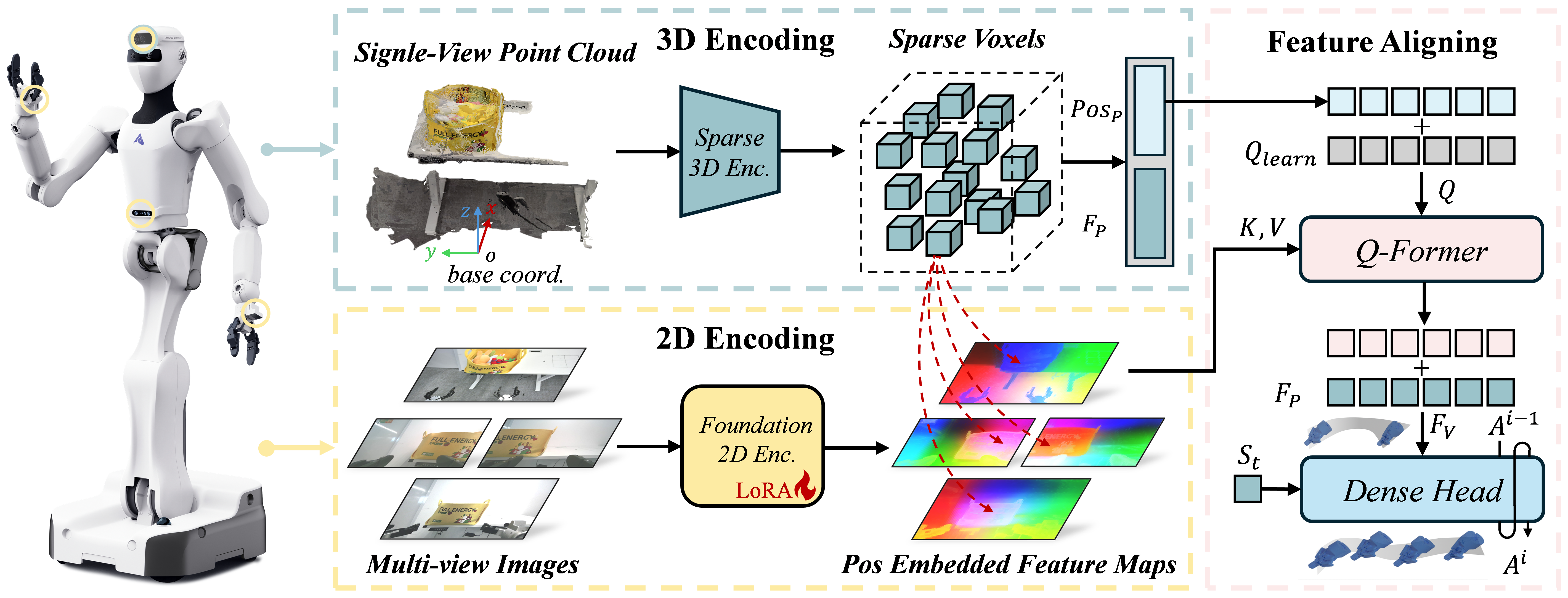}
    \caption{\textbf{Overview of \model}. First, a sparse 3D encoder processes the point cloud, which is projected from the head-mounted camera to the base frame, to obtain voxel-level feature tokens. A 2D vision foundation model is also used to acquire patch-level feature maps. Subsequently, in the feature alignment process, a Q-former is designed to query multi-view semantic features from feature maps for the voxels and fuse them with spatial features, based on the positional information of the 3D voxels and 2D patches. Finally, the resulting features are fed into a dense head to generate the future action sequence in a bidirectional autoregressive paradigm.}
    \label{pipeline}
\end{figure*}
\subsection{Whole-body Mobile Manipulation}
\label{sec:wb}
Although many studies have focused on mobile manipulation~\cite{gamma,acdit,hermes}, work extending to whole-body mobile manipulation remains scarce. Current solutions~\cite{brs} encode the merged point cloud using PointNet~\cite{pointnet} and then generate actions for different robot components, from the chassis to the arms, with diffusion models~\cite{DDPM,DDIM}. The actions for later-generated components are autoregressively conditioned on the former to avoid the error amplification highlighted in this research. Furthermore, this approach has not yet achieved policy-level generalization. Our research just aims to suppress the aforementioned errors and achieve effective manipulation by developing a more efficient visual encoder and an autoregressive policy that leverages temporal dependencies, with an effective method to realize policy-level generalization.

%% file: sec/3_method.tex
\section{Method}

\subsection{Problem Formulation}
Our platform is a 25-DOF robot consisting of a 3-DOF chassis, a 4-DOF torso, dual 7-DOF arms, a 2-DOF head, and 2 grippers. 
The robot is controlled via pose commands for each component relative to the base~\cite{duocorewb}. 
Specifically, the action $\bm{a}_t$ at each timestep $t$ includes the above components' poses, where the chassis's action is an offset from its previous state. 
The task of our policy is therefore to predict the future action sequence $\bm{A}_{t:t+T} = \{\bm{a}_{t}, \bm{a}_{t+1}, \dots, \bm{a}_{t+T-1}\}$ over a horizon of $T$ time steps. 

Given the observation at the current timestep: $\bm{O}_t=\{\bm{I}_t, \bm{P}_t, \bm{S}_t\}$, where $\bm{I}_t$ comprises the RGB images from the robot's head, torso, and dual wrist cameras; $\bm{P}_t$ is the uncolored point cloud from the head camera, projected into the base frame to mitigate the effects of varying head poses on policy learning~\cite{brs}; and $\bm{S}_t$ denotes the current robot state. In line with the standard behavior cloning framework~\cite{bc}, our objective is to learn a policy that maximizes the conditional probability $\bm{P}(\bm{A}|\bm{O})$ over a dataset of expert demonstrations. The overall process of \model is shown in the Figure~\ref{pipeline}

\subsection{Observation Encoding}
For 3D encoding, we use a sparse convolutional network~\cite{spatioenc} to extract voxel features, $\bm{F}_{P}$, from the uncolored geometric point cloud. 
This process, involving convolution and pooling, yields a highly sparse representation containing fewer than 300 tokens in our experiments. 
For each of these feature tokens, we generate a corresponding positional embedding, $\bm{Pos}_{P}$, using a sine-cosine function on its spatial location~\cite{RISE}.

To encode 2D features, we utilize a DINOv2-base backbone~\cite{dinov2} fine-tuned with LoRA~\cite{lora}. This allows the model to adapt to our specific task while retaining its powerful, task-agnostic semantic representations, which is crucial for generalization. This encoder generates patch-level feature maps $\bm{F}^{v_1-v_n}_{I}$ from n views, to which we add learnable positional embeddings, $\bm{Pos}_{I}$, to encode the spatial location of each patch.

The robot state is defined by the current poses of its components, with one important exception: we exclude the chassis's pose from the input. 
Our experiments indicate this forces the policy to rely on environmental context for navigation, rather than its own trajectory history, which improves learning. 
To further prevent the policy from overfitting, we randomly mask the remaining pose inputs with a 30\% probability during training~\cite{dsp}. 
The complete state representation is then encoded into the feature space as $\bm{F}_S$ using an MLP.

\subsection{Feature Aligning}
The visual features subsequently undergo a fusion operation. 
For this, We chose the Q-former~\cite{blip2} architecture, inspired by its success in vision-language modeling, for its effectiveness in distilling salient information from large feature maps into a fixed set of queries, which is ideal for aligning sparse 3D tokens with dense 2D features.

Thus, we design a Q-former that maintains a set of 300 learnable query tokens, $\bm{Q}_{\textit{learn}}$. 
The function of this module is to use the positional information of the sparse spatial tokens to query for corresponding features in the multi-view images, thereby aligning the spatial and semantic features.

Specifically, the learnable tokens are first added to the spatial positional embeddings $\bm{Pos}_P$ to create spatially-aware queries, $\bm{Q}$, as follows:
\begin{equation}
    \bm{Q} = \bm{Q}_{\textit{learn}} + \bm{Pos}_P.
    \label{eq:query_formation}
\end{equation}
These queries then attend to the 2D feature maps from all views ($\bm{F}^{v_1-v_n}_I$), which, augmented with their respective positional embeddings ($\bm{Pos}_{I}$), serve as the keys ($\bm{K}$) and values ($\bm{V}$) in a cross-attention mechanism. 
In this attention process, positional information effectively flows from the 3D voxel grid to the 2D patch grids. 
This flow serves as an efficient indexing mechanism, enabling the policy to align features across multiple views.
The output of this operation, representing the retrieved semantic features, is then added to the original spatial features $\bm{F}_P$. 
This sum forms the final fused visual representation, $\bm{F}_V$, formulated as:
\begin{equation}
    \bm{F}_V = \textit{Attn}(\bm{Q}, \bm{F}^{v_1-v_n}_I + \bm{Pos}_I, \bm{F}^{v_1-v_n}_I + \bm{Pos}_I) + \bm{F}_P.
    \label{eq:final_representation}
\end{equation}

\subsection{Dense Generation}
Beyond the challenges of effective information utilization and generalization, a significant focus in recent research has been the problem of error amplification at the action generation stage~\cite{brs}. 
Specifically, traditional diffusion models are prone to compounding errors, particularly when facing distributional shifts~\cite{d3pm}. 
In the context of diffusion-based policies for whole-body robots, this issue is manifested during the denoising process: an action dimension for one component can only attend to the states of other components at the same noise level. 
For example, during the early denoising steps, the torso's action cannot be conditioned on a fully-denoised state of the mobile base. 
Consequently, coordination between components is compromised, leading to a hierarchical accumulation of errors from the base up to the dual arms~\cite{acdit,brs}.

Although method mentioned in Sec.~\ref{sec:wb} addresses the problem semi-autoregressively, this comes at the cost of substantially prolonged inference latency due to its hierarchical diffusion structure. Such a delay is a considerable drawback in dynamic household environments~\cite{FAST,pi05,rtc,kl}.

Thus, we adopt Dense Policy~\cite{dsp} as our action head, which has been demonstrated to be an inference-efficient autoregressive policy. 
It achieves coarse-to-fine autoregressive generation by predicting the action sequence in a bidirectional, expanding fashion from the observation. 
This characteristic allows the action for one robot component to be conditioned on the already-determined actions of other components at critical timesteps within the trajectory. 
As a result, inter-component coordination is improved, which in turn mitigates the error amplification phenomenon~\cite{ajay2023cond}. 
The logarithmic inference complexity of Dense Policy also reduces the number of inference iterations while increasing the number of causal inference steps in time, thereby curbing the amplification of errors in time series~\cite{fbc}.

As a result, the action generation can be formulated as:
\begin{equation}
    \bm{P}(\bm{A}|\bm{F}_V,\bm{F}_S) = \prod_{i=1}^{n}\bm{P}(\bm{A}^i|\bm{A}^{i-1},\bm{A}^{i-2},\dots,\bm{A}^0,\bm{F}_V,\bm{F}_S),
\end{equation}
where
\begin{equation}
    \bm{A}^n=\{\bm{a}_{t+i}^n|\ i\ mod \ \frac{T}{2^n}= 0,\ \ i \in \mathbb{N}_{<T}\}.
    \label{denselevel}
\end{equation}

%% file: sec/4_exp.tex
\section{Experiments}
\begin{figure*}[!t]
  \centering
    \includegraphics[width=\textwidth]{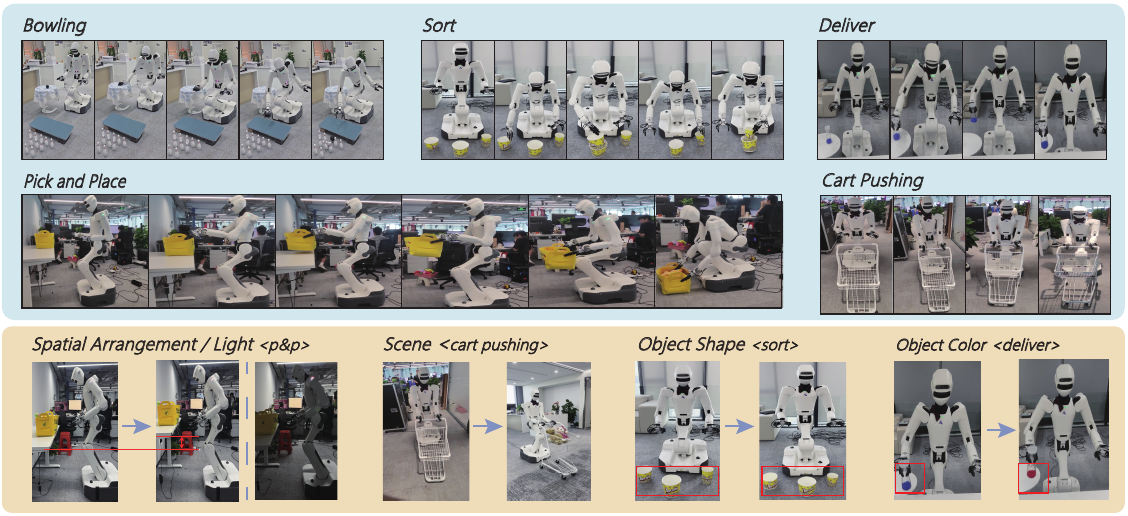}
    \caption{\textbf{Overview of the procedure of our experiments.} The blue area above shows the execution process of the five tasks, and the yellow area below shows the setup of our generalization tests.}
    \label{tasks}
\end{figure*}
\begin{table*}[!t]
    \centering
    \renewcommand\arraystretch{1.6}
    \setlength\tabcolsep{3pt}
    \setlength{\aboverulesep}{0pt}
    \setlength{\belowrulesep}{0pt}
    \small
    \begin{tabular}{c|ccccccccccccc}
        \hline
        \multirow{2}{*}{\textbf{Method}}  & \multicolumn{2}{c}{\textit{\textbf{Pick and Place}}} & \multicolumn{1}{c}{\textit{\textbf{Deliver}}} & \multicolumn{2}{c}{\textit{\textbf{Sort}}} & \multicolumn{2}{c}{\textit{\textbf{Bowling}}}& \multicolumn{1}{c}{\textit{\textbf{Cart Pushing}}} \\
        \cmidrule(r){2-3}\cmidrule{4-4}\cmidrule(l){5-6}\cmidrule(l){7-8}\cmidrule(l){9-9}
        & \textit{Pick (\%)} & \textit{Place (\%)} & \textit{Succ. (\%)} & \textit{Stack-I (\%)} & \textit{Stack-II (\%)} & \textit{Grasp (\%)} & \textit{Hit (\%)} & \textit{Succ. (\%)}  \\
        \hline
        \textit{WB-WIMA}~\cite{brs} & $50$ & $25$ & $45$ & $45$ & $0$ &$65$ &$40$ &$\mathbf{90}$ \\
        \textit{DP3}~\cite{DP3} & $40$ & $10$ & $10$ & $45$ & $10$ &$55$ &$35$ &$80$\\
        \textit{DP}~\cite{DP} & $30$ & $20$ & $55$ & $50$ & $15$ &$50$ &$40$ &$80$\\
        \rowcolor[HTML]{fde8eb}
        \textbf{\textit{\model}} & $\mathbf{80}$ & $\mathbf{60}$ & $\mathbf{100}$ & $\mathbf{90}$ & $\mathbf{25}$ &$\mathbf{80}$ &$\mathbf{50}$ & $\mathbf{90}$ \\ 
        % \textbf{\textit{\model w. Diff Head}}  & $50$ & $30$ & $\mathbf{100}$ & $60$ & $15$ &$\mathbf{90}$ &$35$ &$80$\\
        \hline
    \end{tabular}
    \caption{\textbf{Detailed performance of \model\ and baselines in real-world tasks under the original setup.}}
    \label{realtask}
\end{table*}
\subsection{Tasks}

\subsubsection{Evaluation Tasks}
We evaluate our policy on five distinct tasks:

\textbf{Pick and Place:} The robot navigates to a cloth basket on a table, picks and then places it in an open area on the floor. 

\textbf{Deliver:} The robot grasps a water bottle from a low table and delivers it to the counter. 

\textbf{Sort:} The robot stacks two popcorn buckets scattered on the ground into a largest popcorn tube. We denote the steps of these two stacks as \textit{Stack-I} and \textit{Stack-II}

\textbf{Bowling:} The robot grasps a bowling ball from a table, navigates to a suitable position, and throws the ball to knock down the target pins. The whole process is divided into \textit{Grasp} and \textit{Hit}.

\textbf{Cart Pushing:} The robot must grasp the handle of a shopping cart and push it to navigate to a 1.5 meters distance through an environment while avoiding obstacles.

\subsubsection{Generalization Tests}
Additionally, we designed five challenging generalization tests to evaluate the policy's robustness under conditions not present in the expert demonstrations.

\textbf{Light:} The \textit{Pick and Place} task is performed under low-light conditions, with the main room lights turned off. 

\textbf{Spatial Arrangement:} For the \textit{Pick and Place} task, the height of the table is raised by 5cm. 

\textbf{Object Color:} In the \textit{Deliver} task, the color of the target water bottle is changed to one unseen during training. 

\textbf{Object Shape:} The \textit{Sort} task is performed with popcorn buckets of a novel shape and smaller relative size. 

\textbf{Scene:} The \textit{Cart Pushing} task is executed in a completely new environment with a different layout and background.

The above experiments are shown in the Figure~\ref{tasks}.
\subsection{Setup}
\textbf{Platform.} Our experiments are conducted on the Astribot-S1 robot~\cite{duocorewb}, a platform with 25 degrees of freedom (DOFs). 
For onboard computation, it is equipped with an Intel i7-1370PE CPU. 
The robot's perception system includes an Orbbec Femto Bolt on the head, an Orbbec Gemini 335 on the torso, and an Intel Realsense D401 on each of the two wrists. 
All of these are RGB-D cameras. 
Additionally, the mobile base is mounted with two Livox MID-360 LiDARs, which were not used in our experiments.
All expert demonstrations were collected via teleoperation using a Meta Quest 3S VR headset.

\textbf{Demonstrations.} Unless otherwise specified, we collected 100 demonstrations for each task.

\textbf{Baseline.} In addition to WB-WIMA~\cite{brs}, which also targets whole-body mobile manipulation, we select classical visuomotor policies DP~\cite{DP} and DP3~\cite{DP3} as baselines for comparison. 
Furthermore, to evaluate the effectiveness of Dense Head for whole-body tasks, we also include a variant that pairs our visual encoder with a Diffusion Head for comparison. 

\textbf{Protocols.} For the real-world evaluation, 20 trials are performed per method for each task, unless otherwise specified. All methods are evaluated under closely matched randomized initial scene configurations for each trial.

\begin{table*}[!t]
    \centering
    \renewcommand\arraystretch{1.6}
    \setlength\tabcolsep{3pt}
    \setlength{\aboverulesep}{0pt}
    \setlength{\belowrulesep}{0pt}
    \small
    \begin{tabular}{c|cc cc c c c}
        \hline
        \multirow{3}{*}{\textbf{Method}} & \multicolumn{4}{c}{\textbf{\textit{Pick and Place}}} & \multicolumn{1}{c}{\textbf{\textit{Deliver}}} & \multicolumn{1}{c}{\textbf{\textit{Sort}}} & \multicolumn{1}{c}{\textbf{\textit{Cart Pushing}}} \\
        \cmidrule(r){2-5}\cmidrule(lr){6-6}\cmidrule(lr){7-7}\cmidrule(l){8-8}
        & \multicolumn{2}{c}{\textit{\textbf{Spatial Arrangement}}}  & \multicolumn{2}{c}{\textit{\textbf{Light}}}  & \multicolumn{1}{c}{\textit{\textbf{Object Color}}} & \multicolumn{1}{c}{\textit{\textbf{Object Shape}}}& \multicolumn{1}{c}{\textit{\textbf{Scene}}} \\
        \cmidrule(r){2-3}\cmidrule{4-5}\cmidrule(l){6-6}\cmidrule(l){7-7}\cmidrule(l){8-8}
        & \textit{Pick (\%)} & \textit{Place (\%)} & \textit{Pick (\%)} & \textit{Place (\%)} & \textit{Succ. (\%)} & \textit{Stack-I (\%)} & \textit{Succ. (\%)}  \\
        \hline
        \textit{WB-WIMA}~\cite{brs} & \drop{50}$30$ & \drop{25}$0$ & \drop{50}$30$ & \drop{25}$10$ & \drop{45}$35$ & \drop{45}$20$ & \drop{90}$60$\\
        \textit{DP3}~\cite{DP3} & \drop{40}$0$ & \drop{10}$0$ & \drop{40}$30$ & \drop{10}$0$ & \drop{10}$10$ & \drop{45}$40$ & \drop{80}$60$ \\
        \textit{DP}~\cite{DP} & \drop{30}$30$ & \drop{20}$0$ & \drop{30}$0$ & \drop{20}$0$ & \drop{55}$35$ & \drop{50}$50$ & \drop{80}$60$ \\
        \rowcolor[HTML]{fde8eb}
        \textbf{\textit{\model}} & \drop{80}$\mathbf{50}$ & \drop{60}$\mathbf{20}$ & \drop{80}$\mathbf{60}$ & \drop{60}$\mathbf{40}$ & \drop{100}$\mathbf{85}$ & \drop{90}$\mathbf{80}$ & \drop{90}$\mathbf{80}$ \\ 
        % \textbf{\textit{\model w. Diff Head}}  & \drop{50}$40$ & \drop{30}$0$ & \drop{50}$40$ & \drop{30}$20$ & \drop{100}$\mathbf{90}$ & \drop{60}$40$ & \drop{80}$60$ \\
        \hline
    \end{tabular}
    \caption{\textbf{Detailed performance of \model\ and baselines in Generalization Tests.}
    The left side of the arrow is the performance of the original setup, and the right side is the performance of the generalized setup.}
    \label{generalization}
\end{table*}
\subsection{Effective Manipulation}

As shown in Table~\ref{realtask}, \model achieves the best performance on the majority of tasks. We attribute this result primarily to the following aspects:

\textbf{Effective Environmental Perception.} Compared to baseline policies, \model utilizes and effectively fuses both multi-view semantic features and 3D spatial features. This contributes to a more comprehensive perception of the environment. In the \textit{\textbf{Pick and Place}} task, the robot must first navigate to the target location to pick the basket, and the accuracy of navigation depends on observation and localization of the basket. \model is consistently able to move to a position directly facing the target, an aspect where other policies often deviate. In the \textit{\textbf{Cart Pushing}} task, after grasping the cart, the robot must move 1.5 meters while avoiding obstacles. This requires coordinated actions from the chassis and the arm to adjust its direction, which relies on obstacle detection. The superior spatial perception of \model enables it to achieve the best obstacle avoidance performance.

\textbf{Precise Action Generation.} The comprehensive encoding and fusion of observational information yield finer-grained features for action generation, resulting in higher action precision for \model. This is evident in the \textit{\textbf{Pick and Place}} task, where \model consistently grasps the basket's edge from above. In contrast, other 3D baselines often grasp the side of the basket horizontally, which causes the object to be dropped during the place phase, leading to task failure even after a successful pick. In the \textit{\textbf{Sort}} task, high-precision alignment is required to stack the grasped popcorn bucket into the target bucket. For this, the sparse sampling of 3D baselines is insufficient for accurate depth perception when the two buckets are close, causing errors in the generated actions. Meanwhile, 2D policies have greater difficulty estimating depth. In contrast, \model leverages its efficient feature representation to accomplish this fine-grained action.

\textbf{Coherent Whole-Body Motion.} As previously discussed, a key advantage of Dense Head is that any given robot component can refine its action generation by observing the planned actions of other components within the keyframes in the trajectory. Unlike diffusion models, which are conditioned on noised values, our policy observes raw actions. This allows it to better capture temporal dependencies, thereby mitigating error propagation and enhancing motion coherence.
\begin{table*}[!t]
    \centering
    \renewcommand\arraystretch{1.6}
    \setlength\tabcolsep{3pt}
    \setlength{\aboverulesep}{0pt}
    \setlength{\belowrulesep}{0pt}
    \small
    \begin{tabular}{c|ccccccccccccc}
        \hline
        \multirow{2}{*}{\textbf{Method}}  & \multicolumn{2}{c}{\textit{\textbf{Pick and Place}}} & \multicolumn{1}{c}{\textit{\textbf{Deliver}}} & \multicolumn{2}{c}{\textit{\textbf{Sort}}} & \multicolumn{2}{c}{\textit{\textbf{Bowling}}}& \multicolumn{1}{c}{\textit{\textbf{Cart Pushing}}} \\
        \cmidrule(r){2-3}\cmidrule{4-4}\cmidrule(l){5-6}\cmidrule(l){7-8}\cmidrule(l){9-9}
        & \textit{Pick (\%)} & \textit{Place (\%)} & \textit{Succ. (\%)} & \textit{Stack-I (\%)} & \textit{Stack-II (\%)} & \textit{Grasp (\%)} & \textit{Hit (\%)} & \textit{Succ. (\%)}  \\
        \hline
        \rowcolor[HTML]{fde8eb}
        \textbf{\textit{\model}} & $\mathbf{80}$ & $\mathbf{60}$ & $\mathbf{100}$ & $\mathbf{90}$ & $\mathbf{25}$ &$80$ &$\mathbf{50}$ & $\mathbf{90}$ \\ 
        \textbf{\textit{\model w. Diff Head}}  & $50$ & $30$ & $\mathbf{100}$ & $60$ & $15$ &$\mathbf{90}$ &$35$ &$80$\\
        \hline
    \end{tabular}
    \caption{\textbf{Detailed performance of \model\ and \model with diffusion head in tasks under the original setup.}}
    \label{realtask_ab}
\end{table*}
\begin{table*}[!t]
    \centering
    \renewcommand\arraystretch{1.6}
    \setlength\tabcolsep{3pt}
    \setlength{\aboverulesep}{0pt}
    \setlength{\belowrulesep}{0pt}
    \small
    \begin{tabular}{c|cc cc c c c}
        \hline
        \multirow{3}{*}{\textbf{Method}} & \multicolumn{4}{c}{\textbf{\textit{Pick and Place}}} & \multicolumn{1}{c}{\textbf{\textit{Deliver}}} & \multicolumn{1}{c}{\textbf{\textit{Sort}}} & \multicolumn{1}{c}{\textbf{\textit{Cart Pushing}}} \\
        \cmidrule(r){2-5}\cmidrule(lr){6-6}\cmidrule(lr){7-7}\cmidrule(l){8-8}
        & \multicolumn{2}{c}{\textit{\textbf{Spatial Arrangement}}}  & \multicolumn{2}{c}{\textit{\textbf{Light}}}  & \multicolumn{1}{c}{\textit{\textbf{Object Color}}} & \multicolumn{1}{c}{\textit{\textbf{Object Shape}}}& \multicolumn{1}{c}{\textit{\textbf{Scene}}} \\
        \cmidrule(r){2-3}\cmidrule{4-5}\cmidrule(l){6-6}\cmidrule(l){7-7}\cmidrule(l){8-8}
        & \textit{Pick (\%)} & \textit{Place (\%)} & \textit{Pick (\%)} & \textit{Place (\%)} & \textit{Succ. (\%)} & \textit{Stack-I (\%)} & \textit{Succ. (\%)}  \\
        \hline
        \rowcolor[HTML]{fde8eb}
        \textbf{\textit{\model}} & \drop{80}$\mathbf{50}$ & \drop{60}$\mathbf{20}$ & \drop{80}$\mathbf{60}$ & \drop{60}$\mathbf{40}$ & \drop{100}$85$ & \drop{90}$\mathbf{80}$ & \drop{90}$\mathbf{80}$ \\ 
        \textbf{\textit{\model w. Diff Head}}  & \drop{50}$40$ & \drop{30}$0$ & \drop{50}$40$ & \drop{30}$20$ & \drop{100}$\mathbf{90}$ & \drop{60}$40$ & \drop{80}$60$ \\
        \hline
    \end{tabular}
    \caption{\textbf{Detailed performance of \model\ and \model with diffusion head in Generalization Tests.}
    The left side of the arrow is the performance of the original setup, and the right side is the performance of the generalized setup.}
    \label{generalization_ab}
\end{table*}
For instance, the \textit{\textbf{Deliver}} task requires grasping a bottle from a low table, a maneuver that demands tight coordination among the chassis, torso, and arm. This task admits multiple solutions through different combinations of movements (e.g., the chassis can move further forward, allowing the torso to lean further back), making seamless coordination paramount. The \textit{\textbf{Bowling}} task presents a similar challenge, not only in grasping the ball but also in the release phase. The arm's final release posture is conditioned on both the torso's height and the distance moved by the chassis. Such scenarios place high demands on motion coherence. By leveraging its effective feature processing and bidirectional autoregressive action generation, \model consistently outperforms the baselines in these complex, coordinated tasks.

\subsection{Generalization Ability}

The performance of the policies in the five generalization tests is presented in Table~\ref{generalization}. \model maintains its leading performance over the baselines on the vast majority of tasks, and its relative performance drop compared to the original task setting is minimal. Specifically, its generalization performance is demonstrated in two main aspects: 

\textbf{Spatial Understanding.} In the \textbf{\textit{Spatial Arrangement}} experiment, the table height was increased, while in the \textbf{\textit{Object Shape}} experiment, the sizes of the popcorn buckets were altered. Both modifications significantly change the point cloud occupancy of the scene. Consequently, 3D baselines that rely on sparse sampling for encoding are severely impacted, as these changes lead to substantial shifts in their features after max-pooling, degrading policy performance.

In contrast, \model's Sparse Encoder processes the complete point cloud at the voxel level without information loss, yielding a more robust representation. Furthermore, even when the point cloud shifts, the Q-former's role in binding voxels to their corresponding multi-view semantic features makes the resulting representation more resilient to such disturbances. This leads to more stable observation features and enhances the policy's generalization capability.

It is worth noting that in the second stage of the \textbf{\textit{Sort}} task, where the target bucket's height was substantially reduced, all methods, including \model, failed to perform the pick. This is because the expert demonstrations contained virtually no examples of the right arm picking from such a low height. As established in prior works, current imitation learning policies struggle to generalize to actions that lie significantly outside the distribution of the expert demonstrations.

\textbf{Semantic Understanding.} In the \textbf{\textit{Light}}, \textbf{\textit{Object Color}}, and \textbf{\textit{Scene}} experiments, we altered the ambient lighting, object colors, and scene background, respectively. These modifications introduce significant visual changes, to which the 2D and 3D baselines are particularly sensitive as they lack visual pre-training. In contrast, \model isolates the impact of these visual shifts. Its 3D branch focuses exclusively on spatial information by processing the color-agnostic xyz point cloud. Meanwhile, its 2D branch uses a Foundation Encoder pre-trained on upstream tasks, yielding semantic information with stronger generalization capabilities. By aligning these two feature sources, the model minimizes the performance degradation caused by visual domain shifts.

\subsection{Ablation}
We divide our ablation study into two parts: the Vision Encoder and the Action Head:
\begin{figure}[!h]
  \centering
    \includegraphics[width=0.3\textwidth]{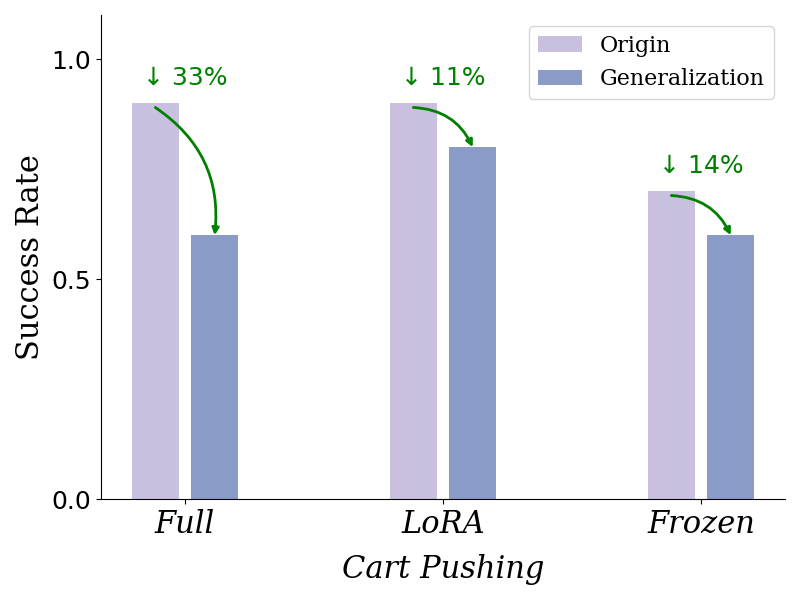}
    \caption{Performance of \model under different fine-tuning strategies for DINOv2 in the \textit{\textbf{Cart Pushing}} task.}
    \label{ablation}
\end{figure}

\textbf{Vision Encoder.} In this section, we investigate the optimal learning method for the 2D vision foundation model to achieve policy-level generalization. We conduct our comparison on the \textit{\textbf{Cart Pushing}} task and its corresponding scene generalization experiment. Specifically, we evaluate three approaches for DINOv2~\cite{dinov2}: full fine-tuning of all parameters, LoRA fine-tuning~\cite{lora}, and freezing all parameters.

The results in Figure~\ref{ablation} show that in the original scene, both the fully fine-tuned and LoRA-tuned models achieve a 90\% success rate, whereas freezing DINOv2 yields only a 70\% success rate. This indicates that LoRA fine-tuning does not sacrifice the policy's precision in extracting semantic features compared to full fine-tuning. In contrast, freezing the vision model prevents it from capturing fine-grained information~\cite{llava}.

When the scene is altered, the success rates change to 60\%, 80\%, and 60\%, respectively. The fully fine-tuned approach suffers a substantial performance drop, as the policy loses the inherent semantic priors of the foundation model. The frozen DINOv2 policy also experiences a larger performance degradation than LoRA. We posit that while freezing the backbone preserves all semantic priors, it prevents the model from extracting sufficient task-level information. This causes the model to rely more heavily on the 3D encoder, which in turn destabilizes the 3D-2D feature binding performed by the Q-former in the generalization setting, reducing the robustness of the visual encoding. Therefore, \model adopts DINOv2 fine-tuned with LoRA as its final 2D encoding method.

\textbf{Action Head.} In Table~\ref{realtask_ab} and \ref{generalization_ab}, we also present the performance of \model when its action head is replaced with a diffusion head. The experimental results show that \model's encoder, when paired with a diffusion head, still outperforms the baselines and exhibits strong generalization. However, the Dense Head demonstrates superior performance.

We attribute this to the mechanism previously discussed: the Dense Policy effectively utilizes keyframe actions from preceding levels within its autoregressive process. This renders the robot's different components mutually ``visible'' during generation, which enhances policy coherence and mitigates error propagation~\cite{dar}. This advantage becomes particularly critical in the generalization tests. When the observation distribution shifts, generative models such as diffusion models are prone to larger initial errors in the observation-to-action mapping~\cite{d3pm,vita}. This error is then significantly amplified during the generation process. In contrast, \model circumvents this problem, leading to more robust performance.

%% file: sec/5_limitations.tex
\section{limitations}
We acknowledge that \model has several limitations.
First, for whole-body mobile manipulation, \model cannot yet solve tasks that require highly constrained or high-frequency actions, which are common in real-world scenarios~\cite{RDP,rtc}; future work could explore incorporating high-frequency modalities like tactile sensing into the policy or addressing this through suitable whole-body action integration.
Second, regarding generalization, \model, like many imitation learning algorithms, fails to generalize to action modes not covered in the expert demonstrations; this might require the design of an action head with stronger reasoning capabilities to provide support.
Finally, a promising direction for household robotics is the development of a general Vision-Language-Action (VLA) model capable of executing diverse, long-horizon, and multi-stage tasks from language instructions. This remains a challenge for future work in whole-body mobile manipulation, and we look forward to the emergence of large-scale whole-body mobile manipulation datasets to support this research, along with VLA designs adapted for them.

%% file: sec/6_conclusion.tex
\section{conclusion}
We propose \model, an efficient and generalizable dense policy for whole-body mobile manipulation. By effectively combining sparse 3D spatial features with multi-view 2D semantic features, \model achieves robust perception of complex environments. Furthermore, we extend the Dense Policy to whole-body tasks, and its bidirectional autoregressive generation mechanism produces coherent and precise actions, significantly outperforming existing methods. Extensive real-world experiments demonstrate that \model achieves significant advancements in both task performance and generalization, offering a promising solution for deploying whole-body robotic systems in complex environments.